%% file: main.tex
\documentclass[]{fairmeta}

\title{SE-GoS: Self-Evolving Graph-of-Skills for Skill Library at Scale}

\renewcommand{\thefootnote}{\fnsymbol{footnote}}
\author[1,2]{Dawei Fu\,\protect\footnotemark[2]}
\author[3]{Cheng Jiang}
\author[4]{Sitian Qian}
\author[1]{Huainan Wang}
\author[5]{Zhongkai Hao}

\affiliation[1]{Tencent}
\affiliation[2]{Peking University}
\affiliation[3]{University of Edinburgh}
\affiliation[4]{Northwestern University}
\affiliation[5]{Tsinghua University}

\usepackage{amsfonts}
\usepackage{amssymb}
\usepackage{booktabs}
\usepackage{algorithm}
\usepackage{algorithmic}
\usepackage{float}
\usepackage{fontawesome5}

\titlespacing*{\paragraph}{0pt}{0.45\baselineskip}{0.5em}

\input{math_commands.tex}

\makeatletter

\renewcommand{\affiliationformat}[2][]{{\small $^{#1}$#2}}
\makeatother

\let\oldcitep\citep
\makeatletter
\renewcommand{\citep}[1]{%
  \begingroup
  \def\hyper@natlinkstart##1{}%
  \def\hyper@natlinkend{}%
  \def\hyper@natlinkbreak##1##2{##1}%
  \def\NAT@hyper@##1{%
    \hyper@linkstart{cite}{cite.\@citeb\@extra@b@citeb}##1\hyper@linkend
  }%
  {\color{metablue}\oldcitep{#1}}%
  \endgroup
}
\makeatother

\abstract{
\input{sections/abstract}
}

\newcommand{\brandicon}[1]{%
  \raisebox{-0.07em}{\includegraphics[height=0.90em]{figures/#1}}}
\newcommand{\githubicon}{\brandicon{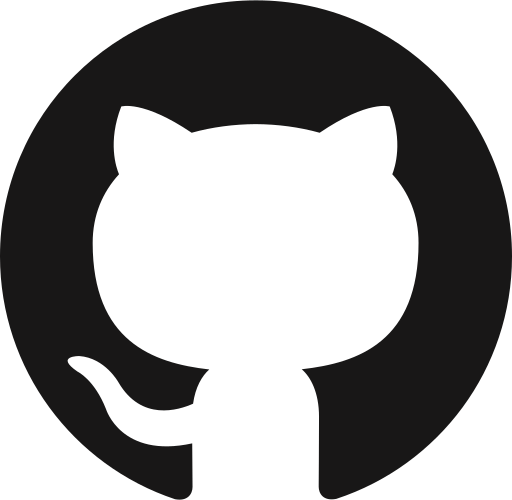}}
\newcommand{\hficon}{\brandicon{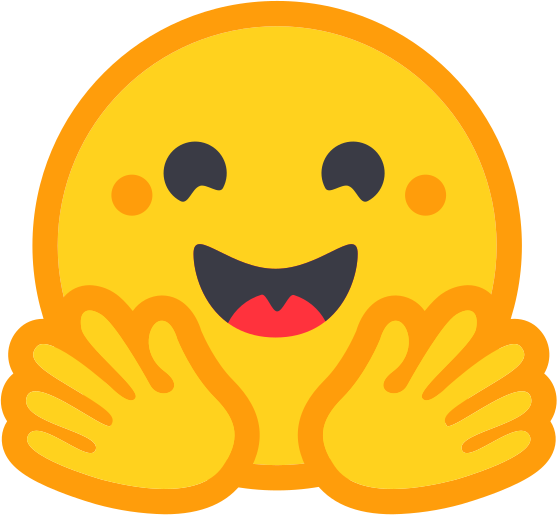}}
\DeclareRobustCommand{\metamark}[1]{\makebox[1.32em][c]{#1}\hspace{0.3em}}
\DeclareRobustCommand{\shorturl}[1]{\href{https://#1}{\texttt{\footnotesize #1}}}
\newcommand{\metaentry}[3]{{\small {\sffamily\bfseries \metamark{#1}#2:} #3}}

\def\metadatalist{%
  \renewcommand{\arraystretch}{0.92}%
  \begin{minipage}{\linewidth}%
  \begin{tabular}{@{}l@{\hspace{0.9em}}l@{}}
    \metaentry{\faClock}{Date}{\today} &
    \metaentry{\scalebox{1.2}{\hficon}}{Data}{\shorturl{huggingface.co/datasets/PKUfudawei/SEGoS-data}} \\
    \metaentry{\faEnvelope}{Correspondence}{Dawei Fu at \email{fudw@pku.edu.cn}} &
    \metaentry{\scalebox{1.2}{\githubicon}}{Code}{\shorturl{github.com/PKUfudawei/SEGoS}} \\
  \end{tabular}%
  \end{minipage}}

\begin{document}

\maketitle

\footnotetext[2]{Work done during an internship at Tencent.}
\renewcommand{\thefootnote}{\arabic{footnote}}
\setcounter{footnote}{0}

\begin{figure}[htbp]
    \centering
    \begin{minipage}[t]{0.58\linewidth}
        \centering
        \vspace{0pt}
        \includegraphics[width=\linewidth]{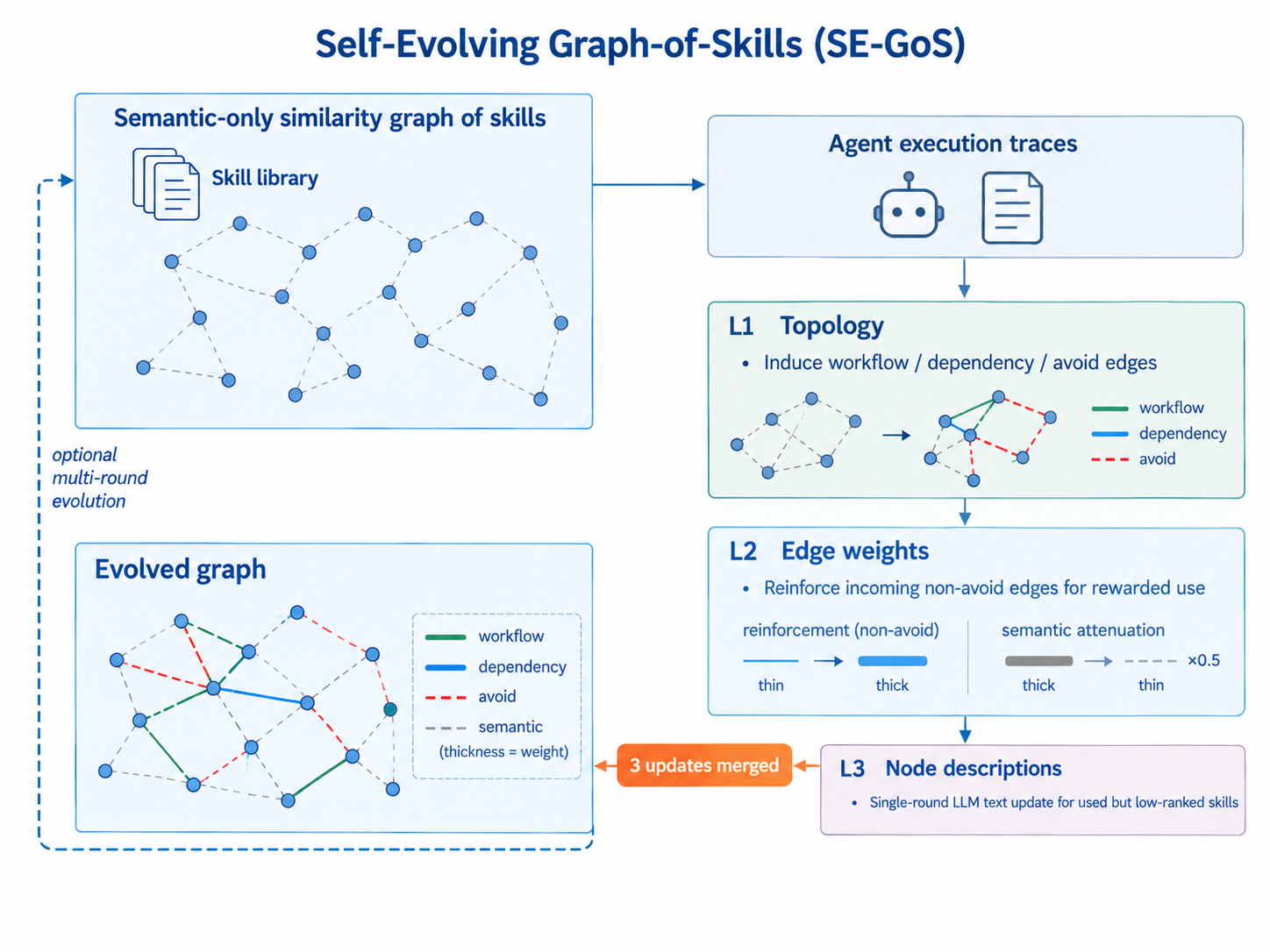}\\[-0.25ex]
        {\small (a) SE-GoS evolution}
    \end{minipage}\hfill
    \begin{minipage}[t]{0.38\linewidth}
        \centering
        \vspace{0pt}
        \includegraphics[width=\linewidth]{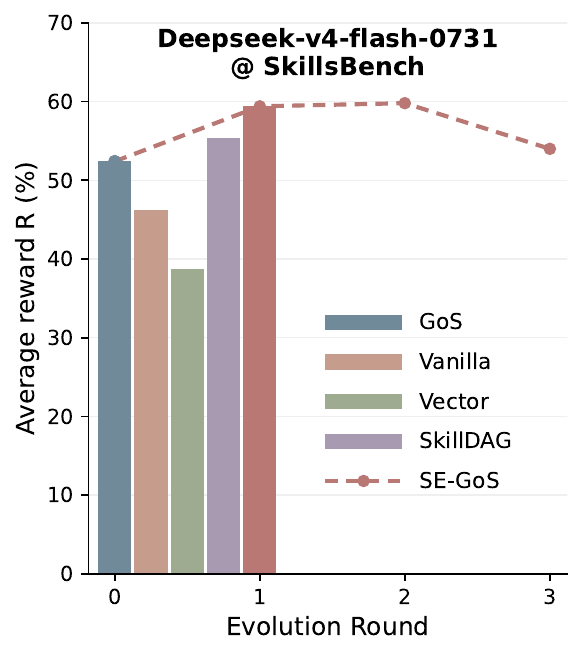}\\[-0.25ex]
        {\small (b) SkillsBench results}
    \end{minipage}
        \caption{SE-GoS overview and SkillsBench results. \textbf{(a)} Offline, trace-driven updates to graph topology, edge weights, and node descriptions; the dashed arrow denotes optional multi-round evolution. \textbf{(b)} SkillsBench reward (\%) for \texttt{deepseek-v4-flash-0731}: bars compare methods; the dashed line shows GoS at round 0 and SE-GoS at rounds 1--3.}
    \label{fig:overview}
\end{figure}

\input{sections/introduction}

\input{sections/background}

\input{sections/method}

\input{sections/experiment}

\input{sections/conclusion}

\bibliographystyle{assets/plainnat}
\bibliography{references}

\input{sections/appendix}

\end{document}

%% file: math_commands.tex
\usepackage{amsmath,amsfonts,bm}

\def\eqref#1{equation~\ref{#1}}

\def\1{\bm{1}}

\DeclareMathAlphabet{\mathsfit}{\encodingdefault}{\sfdefault}{m}{sl}
\SetMathAlphabet{\mathsfit}{bold}{\encodingdefault}{\sfdefault}{bx}{n}



%% file: sections/abstract.tex
LLM agents use large libraries of reusable skills. At thousands of skill entries, retrieval becomes the bottleneck. Graph-of-Skills (GoS) retrieves dependency-aware bundles from a typed skill graph, and SkillDAG shows that such a graph can accumulate execution-backed structure online. Neither asks whether execution traces can be distilled into a better retrieval graph that generalizes to unseen tasks. We present \textbf{Self-Evolving Graph-of-Skills (SE-GoS)}, which treats the retrieval graph as an index rather than a learned representation: the graph is maintained from execution traces while the retrieval pipeline, the skill library, and the model stay fixed. SE-GoS applies three updates: (1) \textbf{topology}, which induces relations from execution evidence and retracts an avoid edge only after repeated successful co-use; (2) \textbf{edge-weight}, which softly attenuates unsupported semantic edges and reinforces incoming edges to used skills; and (3) \textbf{node-description}, which updates retrieval-facing descriptions stored on graph nodes ranked too low. On SkillsBench, one evolution round lifts average reward from 52.4\% to 59.4\%, above full-library loading, vector retrieval, static GoS, and SkillDAG, and this ordering repeats on all three backbones. Retrieval over the evolved graph spends about two-thirds of the input tokens that loading the full library costs. Repeating the round does not help. The same graph improves a held-out split it never saw from 52.9\% to 58.3\%, so what it accumulates transfers rather than memorizes traces. Skill graphs can therefore be improved from execution experience without model training, retrieval-algorithm changes, skill-content modifications, or a model judging which skills are related.

%% file: sections/introduction.tex
\section{Introduction}
\label{sec:intro}

Large Language Model (LLM) agents solve technical tasks by calling external tools and reusable skills~\citep{schick2023toolformer, mialon2023augmented}. As skill repositories grow from dozens to thousands of entries~\citep{patil2023gorilla, li2023apibank, xu2023toolbench, qin2024toolllm}, the bottleneck is no longer whether to use a skill but which small subset to load~\citep{shi2025toolret}. Putting the whole library in context does not scale. Token cost grows linearly, and the model misses skills it needs in an overloaded context~\citep{agentskills2026, liu-etal-2024-lost}. Vector-based retrieval~\citep{lewis2020retrieval,karpukhin2020dense} picks semantically similar skills but ignores their functional prerequisites: the top match is often a high-level solver whose lower-level parser or setup utility is semantically weak yet required for execution. We call this the \emph{prerequisite gap}.

Graph-of-Skills (GoS)~\citep{gos} addresses this with a typed, directed skill graph, seeded by hybrid semantic--lexical signals and explored by reverse-aware Personalized PageRank (PPR)~\citep{page1999pagerank, 10.1145/511446.511513, 10471277}, returning a bounded, budgeted execution bundle. On the 1{,}000-skill SkillsBench under GPT-5.2 Codex, GoS attains a peak reward gain of roughly $25.6\%$ over full loading while cutting token cost by roughly $56.7\%$~\citep{gos}. But the GoS graph is \emph{static}. It is built once, and never updated from execution feedback. When an induced dependency is wrong, a useful relation is missing, or a description systematically fails to match the queries that need it, the repository cannot self-correct.

The signal needed for this is already there. Every trial records which skills were retrieved, which were actually used, and whether the task succeeded. We propose \textbf{Self-Evolving Graph-of-Skills} (SE-GoS), which keeps its lexical-seed reverse-PPR retrieval configuration fixed while improving the graph it reads through three \emph{training-free} updates: topology induction from traces (workflow edges from query-level seed-to-use co-occurrence, dependency edges from ordered usage, avoid edges from failed-trial co-use, and avoid-edge retraction only after repeated successful co-use), edge-weight attenuation of unsupported semantic links and reinforcement toward skills actually used, and single-round textual-gradient optimization of the node descriptions that retrieval surfaces poorly~\citep{protegi}. PPR reads the graph's edge weights directly, so changing the weights or the edges changes what gets retrieved. We do not edit any skill or train a parameter. Experience is therefore stored in the retrieval harness rather than in the model or in the skills.

Our core contributions are as follows:
\begin{enumerate}
    \item We identify the \emph{static-graph bottleneck} of structural skill retrieval, a limitation GoS itself acknowledges~\citep{gos}, and show that the traces a single evaluation already produces are enough to repair it.
    \item We propose SE-GoS, which turns that signal into three training-free updates over the retrieval substrate: execution-grounded relation induction and avoid-edge retraction, soft semantic-edge attenuation and positive edge-weight reinforcement, and single-round node-description optimization. The updates require no model training and do not modify skill files. The node update rewrites only a graph node's retrieval-facing description field, so what changes is the retriever's view of a skill rather than the skill itself.
    \item We explain why editing the graph is enough. PPR reads edge weights directly, so re-wiring the graph changes what gets retrieved, and confidence-weighted updates protect edges with little evidence.
    \item We evaluate in three settings. On the full benchmark one evolution round lifts static GoS from 52.4\% to 59.4\% at 32\% fewer input tokens than full loading, ahead of flat and vector retrieval and of the self-evolving baseline.
    \item We then answer the two questions a static-graph baseline cannot. Repeating the update does not help: the gain plateaus and then reverses, making evolution a one-shot deployment step. The gain also survives on a held-out split the graph never saw (52.9\% to 58.3\%), so what accumulates is transferable structure rather than memorized traces.
\end{enumerate}

%% file: sections/background.tex
\section{Related Work}
\label{sec:relatedwork}

\paragraph{Graph-structured retrieval and skill graphs.}
Graph-structured retrieval improves access to documents and tools~\citep{edge2024graphrag, liu2024toolnet}, often using Personalized PageRank over automatically built graphs, as in HippoRAG~\citep{gutierrez2024hipporag} and tool-graph rerankers~\citep{xu2024toolrerank, yuan2024craft, liu2023controlllm}. GoS~\citep{gos} builds a typed skill graph for dependency-aware retrieval but keeps it static; SkillDAG~\citep{skilldag} edits a typed graph online with LLM proposals, while SkillGraph~\citep{skillgraph-rl} jointly updates teacher-distilled skills and the policy that reads them. Related tool-retrieval work learns collaborative retrieval models, including COLT~\citep{colt2024} and MassTool~\citep{masstool2025}, while Tool-REX studies document expansion and introduces Tool-Embed and Tool-Rank~\citep{toolrex2026}. These works address adjacent problems, but do not evolve a fixed skill-retrieval graph from offline agent traces with the retrieval pipeline, skill library, and backbone held fixed. SE-GoS makes the graph mutable harness state: its deterministic semantic-only substrate is updated from execution evidence, and its node-description update is one component of this graph-evolution framework rather than a standalone document-expansion method. SkillGraph changes the policy and skill representations, so it is not part of our frozen-backbone results.

\begin{table}[htbp]
\centering
\footnotesize
\setlength{\tabcolsep}{3pt}

\begin{tabular}{@{}>{\raggedright\arraybackslash}p{5.5cm}>{\centering\arraybackslash}p{1.35cm}>{\centering\arraybackslash}p{1.3cm}>{\centering\arraybackslash}p{1.35cm}>{\centering\arraybackslash}p{1.7cm}>{\centering\arraybackslash}p{1.25cm}@{}}
\toprule
 \textbf{Method} & \textbf{Graph} & \textbf{Evolving} & \textbf{Training-} & \textbf{LLM-prior-} & \textbf{Skill} \\
 & \textbf{substrate} & & \textbf{free} & \textbf{free} & \textbf{selection} \\
\midrule
Vector retrieval~\citep{karpukhin2020dense} & -- & -- & \checkmark & \checkmark & \checkmark \\
GoS~\citep{gos} & \checkmark & -- & \checkmark & -- & \checkmark \\
SkillRouter~\citep{skillrouter} & -- & -- & -- & \checkmark & \checkmark \\
Voyager~\citep{wang2023voyager} & -- & \checkmark & \checkmark & -- & -- \\
ProTeGi~\citep{protegi} & -- & \checkmark & \checkmark & \checkmark & -- \\
SkillGraph~\citep{skillgraph-rl} & \checkmark & \checkmark & -- & \checkmark & -- \\
SkillDAG~\citep{skilldag} & \checkmark & \checkmark & \checkmark & -- & \checkmark \\
\midrule
\textbf{SE-GoS (ours)} & \checkmark & \checkmark & \checkmark & \checkmark & \checkmark \\
\bottomrule
\end{tabular}
\caption{Positioning of SE-GoS. \emph{Evolving}: retrieval state changes with execution. \emph{Training-free}: no model weights are trained. \emph{LLM-prior-free}: no model judges which skills are related. \emph{Skill selection}: selects among existing skills, without writing them. SkillDAG, the closest row, differs from SE-GoS only on \emph{LLM-prior-free} (Section~\ref{sec:experiment}).}
\label{tab:positioning}
\end{table}

\paragraph{Experience accumulation, prompt optimization, and test-time adaptation.}
A separate line of work already exploits execution traces, but each adapts a different object. Voyager~\citep{wang2023voyager} accumulates a skill library from exploration, and successors distill cross-episode insights, turn failures into verbal reinforcement, synthesize new skills, or induce reusable workflows~\citep{zhao2024expel, reflexion, zheng2025skillweaver, awm}; others compile a skill directory from a document corpus and have the agent browse it~\citep{sun2026distilling}. These systems change the skill library or the agent's procedure. Prompt optimization changes the instruction instead: ProTeGi~\citep{protegi} mirrors gradient descent in language, and TextGrad~\citep{textgrad} formalizes the same principle as automatic differentiation over textual artifacts. Test-time training~\citep{sun2020ttt} changes the model itself, adapting \emph{weights} to the test instance. SE-GoS leaves all three objects untouched and restructures only the graph that retrieval traverses: its node update borrows ProTeGi's machinery for a single-round, node-local variant aimed at used-but-low-ranked skills, but the skill library, the agent's procedure, and the model weights all stay as they were.

\paragraph{Positioning.}
The distinguishing combination is structural, dependency-aware retrieval over a skill graph that evolves from execution traces, with no model trained, no LLM prior over which skills are related, and no modification of skill content or retrieval code. Table~\ref{tab:positioning} compares selected skill-selection and graph-evolution paradigms; adjacent tool-retrieval work is discussed above but is not a direct baseline in our experiments.

%% file: sections/method.tex
\section{Method: SE-GoS}
\label{sec:method}

SE-GoS lets the GoS retrieval procedure read a substrate that \emph{evolves} from execution feedback (Figure~\ref{fig:overview}). The substrate is evolvable without code changes because retrieval reads its edge weights and node-level retrieval descriptions: re-weighting or re-wiring the graph, or rewriting a node description, changes what is retrieved.

The cold-start graph is not GoS's typed graph, since we do not assume the LLM validation pass that builds its workflow and semantic relations. SE-GoS begins from a deterministic semantic-only similarity graph, with no LLM pass and no embedding service, and execution then supplies the missing structure, replacing GoS's LLM prior with observed experience. Each graph node has its own retrieval-description field, initialized from the associated skill record; the node-description update changes only this graph-local field, not the source SKILL.md, package metadata, or hydrated skill content.

\subsection{Preliminaries}
\label{sec:prelim}
SE-GoS is built on the Graph-of-Skills (GoS) retrieval substrate~\citep{gos}, restated here as designed by GoS and inherited without modification. The configuration used in our experiments is reported in Section~\ref{sec:experiment}.

\paragraph{Problem setup.}
Let $\mathcal{C}=\{d_1,\dots,d_m\}$ denote a local corpus of skill packages. Following GoS, each skill is normalized into an executable record and the corpus becomes a typed directed graph
\begin{equation}
G = (V, E, w, \phi),
\label{eq:graph}
\end{equation}
where each node $v \in V$ is a normalized skill with a graph-local retrieval description $d_v$, each edge $e \in E$ connects two skills, $w(e) \ge 0$ is an edge weight, and $\phi(e) \in \mathcal{R}$ assigns an edge type from the relation set
\begin{equation}
\mathcal{R} = \{\text{semantic}, \text{workflow}, \text{dependency}, \text{avoid}\},
\label{eq:relations}
\end{equation}
\emph{Semantic} is the relation our substrate starts from, and it is a deterministic token-overlap similarity graph rather than GoS's LLM-validated semantic edges. \emph{Workflow} and \emph{dependency} are GoS's remaining navigation relations, dependency induced deterministically by matching producer outputs against consumer inputs and workflow obtained by a sparse \emph{LLM relation validator} over a bounded candidate pool (validation budget $k{=}8$ per node, Section~\ref{sec:appendix_hyper}). The graph GoS builds therefore carries an LLM-supplied prior about inter-skill structure, constructed offline before any agent runs. GoS's fourth navigation relation, \emph{alternative}, is one SE-GoS does not use, since whether one skill can stand in for another is a claim about an execution that did not occur (Appendix~\ref{sec:appendix_design}). \emph{Avoid} is the relation SE-GoS adds: it marks a pair co-used on failed traces and not co-used on successful traces, carries weight zero, and is never traversed, so it adds no diffusion mass and affects retrieval only at bundle composition (Section~\ref{sec:evolution}).

Given a query $q$ and context budget $\tau$, retrieval returns a bounded, execution-complete bundle $B(q) \subseteq V$.

\paragraph{Lexical seeding.}
We use GoS's lexical-seed evaluation condition, so the $i$-th component of the seed distribution is $p_i \propto s_i^{\text{lexical}}(q)$ and no embedding service is required at inference. The local scorer uses weighted token overlap over the skill name, capability, description, domain tags, tooling, I/O types, example tasks, and entrypoints.

\paragraph{Reverse-aware typed diffusion.}
Let $A_r$ denote the weighted adjacency matrix for relation type $r \in \mathcal{R}$, with entries drawn from the edge weights $w(\cdot)$. For each type, GoS forms a row-normalized forward operator $T_r^{\rightarrow}$ and a row-normalized reverse operator $T_r^{\leftarrow}$, and unifies them into
\begin{equation}
T = \operatorname{RowNorm}\!\left(A^{\rightarrow} + A^{\leftarrow}\right),
\label{eq:transition}
\end{equation}
where $A^{\rightarrow}_{uv}$ accumulates the stored weight $w(u\to v)$ of every edge from $u$ to $v$, and $A^{\leftarrow}_{vu}$ accumulates $\gamma_{\phi(u,v)}w(u\to v)$. The reverse coefficients are fixed by relation type; no separate forward relation multiplier is used. An avoid edge carries weight zero and so contributes nothing to $T$. Retrieval then runs reverse-aware Personalized PageRank diffusion~\citep{page1999pagerank, jeh2003scaling, 10471277},
\begin{equation}
\mathbf{s}^{(\ell+1)} = \alpha \, \mathbf{p} + (1-\alpha) T^{\top} \mathbf{s}^{(\ell)},
\label{eq:ppr}
\end{equation}
with restart probability $\alpha \in (0,1)$, so that relevance propagates from the matched seeds toward structurally important prerequisites.

\paragraph{Budgeted hydration.}
The lexical field matches enter at seeding time; after PPR, skills are ranked by $\mathbf{s}^{\star}$. The top-ranked skills are hydrated under per-skill and global context budgets, yielding a bounded execution bundle.

\subsection{Experience Signal Collection}
\label{sec:signal}
SE-GoS consumes artifacts a single GoS evaluation already produces, with no additional runs. Each trial $t$ on task $q_t$ yields a trace
\begin{equation}
\mathcal{T}_t = \big( q_t,\; r_t,\; B_t,\; U_t,\; \text{tokens}_t \big),
\label{eq:trace}
\end{equation}
where $r_t \in [0,1]$ is the verifier reward, $B_t \subseteq V$ is the retrieved bundle, and $U_t$ is the set of skills the agent \emph{actually used}, extracted from the trajectory's tool calls by matching the hydrated skills' local source paths. $U_t$ is unordered and is used for co-use evidence; when temporal order is needed for dependency induction, we separately retain the first-use sequence $U_t^{\mathrm{ord}}$. $U_t$ may include skills outside $B_t$ (an agent can discover a skill by browsing the library even when retrieval misses it), the case the node update targets. We denote the traces over an evolution window as $\mathcal{D} = \{\mathcal{T}_1, \dots, \mathcal{T}_T\}$.

The three updates act on the retrieval substrate along three complementary axes (the discrete edge set, the continuous edge weights, and the node description), so that no update rewrites another's output.

\subsection{Topology Update}
\label{sec:evolution}
The topology update changes the graph's discrete structure by inducing workflow, dependency, and avoid relations. It can also retract an existing avoid edge when repeated successful co-use directly contradicts its failure-side evidence. Non-use alone never deletes an edge; semantic-edge attenuation belongs to the edge-weight update. The induced relations differ in their weights as well as in their evidence,
\begin{equation}
w(u \to v) =
\begin{cases}
\min\bigl(0.9,\; 0.6 + 0.05\,(c^{r}_{uv} - 1)\bigr), & r\in\{\text{workflow},\text{dependency}\},\\[2pt]
0, & \text{avoid edges},
\end{cases}
\label{eq:induce}
\end{equation}
where $c^{\text{workflow}}_{uv}$ counts retrieval queries that witness the pair (so one task with multiple queries can contribute more than once), while $c^{\text{dependency}}_{uv}$ counts distinct task traces that witness the ordered pair.

\paragraph{Workflow edges.}
A seed skill $u$ retrieved for a query in a successful trial and a different skill $v$ that the agent uses are a query-level behavioral co-occurrence. Each witnessing query increments $c^{\text{workflow}}_{uv}$, so a task with multiple queries may contribute more than once. The trace does not establish that retrieving $u$ caused the later use of $v$.

\paragraph{Dependency edges.}
When a successful trial uses $u$ before $v$ and the schema overlap between $u$'s outputs and $v$'s inputs clears the threshold GoS applies offline ($\zeta{=}0.6$), the schema predicts the prerequisite and the trajectory confirms it. This requires the trace to record skill order, which the full-benchmark protocol of Section~\ref{sec:experiment} logs.

\paragraph{Avoid edges.}
A pair of skills used together on \emph{failed} trials at least $\theta_{\text{avoid}}$ times and never together on a successful one is evidence for an \emph{avoid} edge. This signal is based on observed co-use, not merely co-loading in a retrieved bundle. We enforce a conservative \emph{non-contradiction} invariant: an unordered pair that already has any navigation relation, including a semantic edge, does not also receive an avoid edge. This can miss harmful pairs among semantically similar skills, but avoids placing opposing signals on the same pair. Avoid edges carry zero weight, are excluded from weight reinforcement, and act only during bundle composition. A candidate whose avoid partner is already selected is dropped without backfilling from below the pre-ranked top-$N$ list, so the final bundle can be smaller than $N$. An existing avoid edge is retracted only if the pair is used together in at least two distinct successful task traces in the current evolution window; simple non-use is not evidence for retraction. What this channel yields on the benchmark is reported in Appendix~\ref{sec:appendix_avoid}.

\subsection{Edge-Weight Update}
This update operates on continuous edge weights, complementing L1's relation-level topology changes. First, if a skill $v$ is surfaced by at least $\theta_{\text{obs}}$ distinct training tasks and is never used in any training trace, each incoming semantic edge is attenuated once, $w(u \to v) \leftarrow 0.5\,w(u \to v)$. This is soft pruning: no edge is deleted and no floor is applied. Multiple queries within one task count as one surfaced task. Second, for each trial $t$ with positive reward ($r_t > 0$) and each used skill $v \in U_t$, every existing non-avoid edge pointing into $v$ is reinforced, regardless of whether its source was retrieved,
\begin{equation}
w^{(t+1)}(u \to v) = w^{(t)}(u \to v) + \eta \cdot r_t \cdot \mathbb{I}[v \in U_t],
\label{eq:reinforce}
\end{equation}
where $\eta > 0$ is the reinforcement rate and $r_t$ scales the increment by task reward; failed trials contribute no positive credit. This update is global over the target's incoming non-avoid edges, not a causal attribution to a retrieved source. We report these raw accumulated weights directly in the single-round protocol evaluated here. Appendix~\ref{sec:appendix_design} derives a confidence-weighted interpolation with the static prior, which guards a deployed graph against cold start in multi-round evolution.

\subsection{Node Update}
\label{sec:method_node}
This update targets \emph{used but poorly surfaced} skills: a skill $s$ used on a failed trial ($r_t < r_{\text{success}}{=}0.9$) but ranked below $n_{\text{rank}}$ (default $3$; an absent skill ranks last) for query $q$. Adapted from ProTeGi~\citep{protegi}, it uses a single-round, node-local textual-gradient operator: an LLM proposes a rewrite and paraphrases, then offline retrieval selects the node description that best improves the target node's rank on these miss queries. Let $H_s(d)$ count miss queries where $s$ appears in the retrieved top-$N$, and $R_s(d)$ sum its 1-indexed ranks on those queries. The selected node description is
\begin{equation}
d_s^{\star} \;=\; \operatorname*{arg\,max}_{d \in \operatorname{Cands}(s)} \; \bigl(H_s(d), -R_s(d)\bigr)_{\mathrm{lex}},
\label{eq:desc}
\end{equation}
where an absent skill has rank $N{+}1$. Candidates must place $s$ within rank $n_{\text{rank}}$ on at least one miss query. The objective first maximizes $H_s(d)$, then minimizes $R_s(d)$; candidate generation, tie handling, and the no-eviction guard are detailed in Appendix~\ref{sec:appendix_hyper}. Candidate selection uses offline retrieval only and requires no additional agent trials.

The model is an \emph{operator} on observed traces, not a \emph{prior} about skill relatedness. Use on a failed task is not a ground-truth relevance label: the agent may have selected an inappropriate skill. The update therefore optimizes alignment with observed use, not verified causal usefulness; the no-eviction guard limits collateral retrieval regressions but cannot validate a positive association. The same trace-conditioned operator is used in all components, keeping the ablation attributable to evolution rather than to a baseline prior.

\subsection{Evolved-Graph Retrieval}
\label{sec:retrieval}
SE-GoS evolves the graph once per evolution window and then serves every query by lexical seeding, reverse-aware PPR, and budgeted hydration over the evolved graph $G^{(T)}$, weights $w^{(T)}$, and descriptions $d^{\star}$. The ranking is therefore \emph{experience-aware}: it depends on graph structure, query relevance, \emph{and} the historical effectiveness encoded in the weights. Evolution is a two-phase, fully offline procedure: Algorithms~\ref{alg:extract} (Phase~A: signal extraction) and~\ref{alg:update} (Phase~B: offline update application) in Appendix~\ref{sec:appendix_alg} spell it out, and Algorithm~\ref{alg:retrieve} (same appendix) shows that retrieval reads the evolved graph.

%% file: sections/experiment.tex
\section{Experiments}
\label{sec:experiment}

Experiments run on SkillsBench~\citep{li2026skillsbench} and the ALFWorld dev split. The baselines are Vanilla, which loads the full library into context, Vector retrieval, static GoS, and SkillDAG~\citep{skilldag}, the closest self-evolving system. We report three studies. Table~\ref{tab:main} compares all five methods on three backbones: two from GoS and one recent cost-effective model that we add. The two follow-up studies run on SkillsBench alone, which retains headroom; ALFWorld is close to saturation on two of the three backbones, where the graph-based rows already sit in the high $80$s to low $90$s (Table~\ref{tab:main}). Section~\ref{sec:heldout} evolves the graph on 50 tasks and measures it on a disjoint 37, which separates transfer from memorization. Section~\ref{sec:multiround} runs further evolution rounds on the settled graph.

\subsection{Experimental Setup}

\paragraph{Benchmark and backbones.}
We evaluate on \textbf{SkillsBench}~\citep{li2026skillsbench}, the benchmark used by GoS~\citep{gos}: real-world technical tasks paired with curated skills. We use the full 87-task set and the released 1{,}000-skill library, and additionally run the ALFWorld \texttt{valid\_seen} split~\citep{shridhar2020alfworld} under the same protocol as a cross-domain check. ALFWorld is an in-domain cross-domain evaluation here; we do not claim held-out ALFWorld transfer. Table~\ref{tab:main} covers three backbones: two are the ones GoS reports, and the third is \texttt{deepseek-v4-flash-0731}, a recent cost-effective flash model that we add as the primary setting. Each setting runs two attempts per task, with up to eight concurrent Docker trials under the GoS environment and retry policy. For our SkillsBench runs, the verifier reward is the fraction of collected unit-test cases passed, computed from the Common Test Report Format (CTRF) output; thus \textbf{R} is the mean of per-attempt rewards in $[0,1]$. This fractional local metric differs from SkillsBench's upstream all-tests-pass binary reward. Since each attempt can receive fractional credit, the pooled mean is not restricted to multiples of $1/174$. ALFWorld reward is binary success, so its \textbf{R} is a success rate. We also report average input tokens per attempt (\textbf{T}) and agent-only runtime (\textbf{S}). Appendix~\ref{sec:appendix_setup} documents the benchmark and backbone choices and the attempt accounting.

\paragraph{Evaluation protocol.}
\label{sec:protocol}
Self-evolving retrieval re-measures the tasks that generated its experience. We follow SkillDAG's deployment-utility convention on SkillsBench: all $87$ tasks produce the traces, the three updates run once, and all $87$ tasks are re-measured on the evolved graph. This is the protocol under which a deployed system is worth measuring, and it puts our full-benchmark cells on the same footing as the published SkillsBench numbers. Its cost is possible memorization of the traced tasks, which Section~\ref{sec:heldout} measures directly. Every pooled reward is computed over \emph{scored} attempts: an attempt in which the harness itself fails before the verifier can run is excluded from both the numerator and the denominator rather than scored as a zero, following the accounting convention of the closest baselines~\citep{skilldag}. Every full-benchmark cell is measured at its full $87\times2=174$ scored attempts. For evolution, the signal parser retains the final attempt of each task, yielding $87$ per-task traces; it does not pool attempts by default. Each number is a mean over scored attempts. We report point estimates descriptively and do not test individual cells for significance; narrow differences should not be interpreted as established statistically significant gains. In the two commercial-backbone blocks, Vanilla, Vector, and static GoS are historical values quoted from GoS, whereas SE-GoS and SkillDAG are fresh measurements, so possible API-version drift is a limitation of these cross-time comparisons. Concurrent reruns and uncertainty estimates are deferred to a follow-up evaluation.

\paragraph{Configuration and evolution protocol.}
The main-table GoS baseline uses the original typed GoS graph; it is distinct from SE-GoS's semantic-only cold-start graph. SE-GoS uses lexical seeding and reverse-aware PPR (top-$N{=}5$/seed-$K{=}4$, 1{,}800-character per-skill / 9{,}000-character global budget) over a deterministic token-overlap $k{=}1$ graph on the official 1{,}000-skill nodes (863 semantic edges). Its evolution traces are collected on this semantic-only substrate, and L1--L3 are applied once offline in dependency order. The component ablation uses the same SE-GoS cold-start graph in all eight conditions, isolating the updates on that substrate; it is not initialized from the original GoS typed graph. The induced relations are execution-grounded (workflow from retrieval-to-use co-occurrence, dependency where ordered usage matches GoS's I/O-schema rule, and avoid from failure co-occurrence), with counts reported alongside the results. Further protocol details are in Appendix~\ref{sec:appendix_alg}.

\subsection{Main Results}

\begin{table}[htbp]
\caption{\textbf{Main results.} \textbf{R}: average reward (\%); \textbf{T}: average input tokens per attempt (M); \textbf{S}: agent-only runtime (s), one decimal. \textbf{Bold} = best, \underline{underline} = second-best per metric column within each model block; $\uparrow$/$\downarrow$ indicate larger/smaller is better. SkillsBench covers all $87\times2=174$ attempts per cell; ALFWorld \texttt{valid\_seen} covers all $140\times2=280$ attempts per block (Section~\ref{sec:protocol}).}
\label{tab:main}
\centering
\small
\setlength{\tabcolsep}{9pt}
\begin{tabular}{@{}p{4.6cm} l ccc ccc@{}}
\toprule
 & & \multicolumn{3}{c}{SkillsBench} & \multicolumn{3}{c}{ALFWorld}\\
\cmidrule(lr){3-5}\cmidrule(lr){6-8}
Model & Method & R $\uparrow$ & T $\downarrow$ & S $\downarrow$ & R $\uparrow$ & T $\downarrow$ & S $\downarrow$ \\
\midrule
\multirow{5}{*}{\texttt{deepseek-v4-flash-0731}}
 & Vanilla       & 46.2 & 5.06 & \textbf{771.4} & 80.4 & 1.81 & 85.1 \\
 & Vector        & 38.7    & \textbf{3.11} & \underline{790.7} & 83.9 & \textbf{0.04} & \textbf{60.3} \\
 & GoS           & 52.4 & 3.67 & 843.8 & 88.2 & 0.07 & 65.2 \\
 & SkillDAG  & \underline{55.3}  & 3.62 & 862.9 & \underline{90.4} & \underline{0.06} & 66.8 \\
 & SE-GoS        & \textbf{59.4} & \underline{3.45} & 883.7 & \textbf{91.1} & 0.07 & \underline{63.2} \\
\midrule
\multirow{5}{*}{\texttt{minimax-m2.7}}
 & Vanilla       & 17.2 & 0.94 & 580.7 & 47.1 & 2.18 & 88.6 \\
 & Vector        & 10.4    & \textbf{0.85} & \underline{552.9} & 50.7 & \underline{0.07} & \underline{73.4} \\
 & GoS           & 18.7 & \underline{0.87} & \textbf{502.5} & 54.3 & \textbf{0.07} & \textbf{68.8} \\
 & SkillDAG  & \underline{27.3} & 1.05 & 560.2 & \underline{67.1} & 0.09 & 75.9 \\
 & SE-GoS        & \textbf{28.5} & 0.92 & 572.4 & \textbf{68.6} & 0.08 & 74.3 \\
\midrule
\multirow{5}{*}{\texttt{gpt-5.2-codex}}
 & Vanilla       & 27.4 & 3.19 & \textbf{686.8} & 89.3 & 1.44 & 83.3 \\
 & Vector        & 21.5 & \textbf{1.24} & 773.0 & \underline{92.9} & \textbf{0.03} & \textbf{57.0} \\
 & GoS           & 34.4 & \underline{1.38} & \underline{715.6} & \textbf{93.6} & 0.05 & 64.7 \\
 & SkillDAG  & \underline{36.8} & 1.65 & 780.4 & \textbf{93.6} & 0.05 & 64.2 \\
 & SE-GoS        & \textbf{38.1} & 1.56 & 749.1 & \textbf{93.6} & \underline{0.05} & \underline{62.3} \\
\bottomrule
\end{tabular}
\end{table}

On \texttt{deepseek-v4-flash-0731}, the original GoS graph scores 52.4\%, versus 46.2\% for Vanilla and 38.7\% for Vector. One SE-GoS evolution round reaches \textbf{59.4\%} ($+7.0$ over GoS), exceeding SkillDAG (55.3\%) and both flat baselines with 3.45M input tokens per attempt ($32\%$ below Vanilla). The component ablation in Section~\ref{sec:ablation} instead starts from SE-GoS's semantic-only cold-start graph, isolating the three updates without testing evolution from the original typed GoS graph. Across the other backbones, SE-GoS improves over GoS by $+9.8$ on \texttt{minimax-m2.7} and $+3.7$ on \texttt{gpt-5.2-codex}; the gain is largest where the static graph scores lowest ($18.7\%$). SkillDAG~\citep{skilldag} reports the same pattern for its routing signal.

\subsection{Generalization to Held-Out Tasks}
\label{sec:heldout}
The reward gain survives on tasks the graph never saw. To separate generalization from memorization, we split the tasks into 50 training and 37 evaluation tasks, disjoint but stratified across the eight skill domains (\texttt{evo\_data/split.json}). Evolution runs on the 50 training tasks only. Every method is then measured on the same 37 evaluation tasks, two attempts each on the $k{=}1$ substrate ($n{=}74$, Table~\ref{tab:heldout}).
\begin{table}[h]
\centering
\footnotesize
\caption{\textbf{Held-out comparison} on the 37-task held-out split ($n{=}74$ attempts, all scored; backbone: \texttt{deepseek-v4-flash-0731}). \textbf{Bold} = best, \underline{underline} = second-best; every row is measured on this split.}
\label{tab:heldout}
\begin{tabular}{lccc}
\toprule
Condition & R $\uparrow$ & T $\downarrow$ & S $\downarrow$ \\
\midrule
Vanilla & 44.8 & 5.23 & \textbf{608.9} \\
Vector & 38.1 & \textbf{3.01} & 573.4 \\
GoS & 52.9 & 3.30 & \underline{631.4} \\
SkillDAG (evolved on train-set) & \underline{54.2} & 3.65 & 721.7 \\
SE-GoS (evolved on train-set)   & \textbf{58.3} & \underline{3.19} & 650.8 \\
\bottomrule
\end{tabular}
\end{table}

On the held-out tasks SE-GoS lifts static GoS from 52.9\% to 58.3\% ($+5.4$ points) with tokens essentially unchanged ($3.30\!\to\!3.19$\,M per attempt) and runtime comparable ($631.4\!\to\!650.8$\,s). Because the edges encode cross-task regularities between skills rather than anything specific to one trace, and the evaluation tasks are disjoint from training, the gain reflects transferable structure rather than memorized traces. As on the full benchmark, the graph-based rows sit above the flat rows, so the ordering is stable across both protocols. SkillDAG~\citep{skilldag} publishes no held-out SkillsBench number (its only held-out result is on ALFWorld), so the SkillDAG row above is measured on this split rather than quoted.

\subsection{Does Multi-Round Evolution Help?}
\label{sec:multiround}
It helps once and then stops. Table~\ref{tab:multiround} re-runs all three updates on the settled graph, with the previous round's traces as the new experience, and re-measures all $87$ tasks each round ($n{=}174$; backbone: \texttt{deepseek-v4-flash-0731}); round~0 is the static graph and round~1 is the SE-GoS cell of Table~\ref{tab:main}.

\begin{table}[h]
\centering
\footnotesize
\caption{\textbf{Multi-round evolution} ($n{=}174$ attempts per round, all scored; backbone: \texttt{deepseek-v4-flash-0731}). \textbf{Bold} = best / \underline{underline} = second-best per metric column.}
\label{tab:multiround}
\begin{tabular}{lccccc}
\toprule
Evolution round & R $\uparrow$ & T $\downarrow$ & S $\downarrow$ & Edges & Node edits \\
\midrule
0 (static)        & 52.4 & 3.67 & 843.8 & 863 & -- \\
1                 & \underline{59.4} & \textbf{3.45} & 883.7 & 1{,}118 & 6 \\
2                 & \textbf{59.8} & \underline{3.66} & \textbf{813.9} & 1{,}375 & 2 \\
3                 & 54.0 & 3.71 & \underline{850.8} & 1{,}502 & 4 \\
\bottomrule
\end{tabular}
\end{table}

Reward rises to $59.4\%$ at round~1, is essentially flat at round~2 ($59.8\%$), then drops to $54.0\%$ at round~3 as edges grow from $1{,}118$ to $1{,}502$. Token cost drifts up as well, and runtime is lowest at round~2 ($813.9$\,s). The edges induced after the first round stop paying for themselves, so we apply the update once at deployment, the setting of the main text.

\subsection{Component Ablations}
\label{sec:ablation}
All eight factorial conditions start from the same SE-GoS semantic-only cold-start graph, not the original typed GoS graph used in the main results. Edge-weight updates have the largest single-component gain (+6.5 points over the 50.8\% control), followed by topology (+3.3) and node descriptions (+2.8). Factorial main effects are +5.1 points for edge weights, +1.8 for topology, and +1.7 for node descriptions; leave-one-out comparisons give the same ordering. The ablation isolates the three updates on SE-GoS's own cold-start substrate; it does not test evolution initialized from the original GoS graph. Additional protocol details are in Appendix~\ref{sec:appendix_setup}.

\begin{table}[htbp]
\centering
\footnotesize
\caption{\textbf{Component ablation} from the SE-GoS semantic-only cold-start graph ($k{=}1$, $n{=}174$ per cell). \textbf{Bold} = best / \underline{underline} = second-best per column.}
\label{tab:ablation}
\begin{tabular}{lccc}
\toprule
Method & R $\uparrow$ & T $\downarrow$ & S $\downarrow$ \\
\midrule
SE-GoS cold-start                       & 50.8 & 3.74 & 829.1 \\
$+$Topology                         & 54.1 & 3.37 & \underline{781.8} \\
$+$Edge                             & 57.3 & \underline{3.35} & 807.9 \\
$+$Node                             & 53.6 & 3.86 & 883.3 \\
$+$Topology$+$Edge                  & 58.8 & 3.66 & 799.4 \\
$+$Topology$+$Node                  & 55.8 & \textbf{3.17} & 844.5 \\
$+$Edge$+$Node                      & \underline{59.1} & 3.43 & \textbf{771.5} \\
$+$Topology$+$Edge$+$Node (SE-GoS) & \textbf{59.4} & 3.45 & 883.7 \\
\bottomrule
\end{tabular}
\end{table}

Adding node updates to topology+edge raises reward from 58.8\% to 59.4\%; this 0.6-point difference is descriptive, and interaction effects are not tested for significance.

%% file: sections/conclusion.tex
\section{Conclusion}
\label{sec:conclusion}

SE-GoS makes execution experience reusable by treating the retrieval graph as offline-maintained harness state, while leaving the retrieval pipeline, backbone, and source skill packages fixed. In the evaluated settings, one evolution round improves reward and transfers to held-out SkillsBench tasks. The component ablation identifies edge-weight adaptation as the largest standalone contributor on the semantic-only cold-start graph, while repeated updates do not yield consistent gains. Unlike online graph-editing approaches, SE-GoS separates experience collection from graph maintenance, so updates can be inspected before they affect subsequent retrieval; it therefore does not adapt immediately within a deployment run. This positions SE-GoS as a trace-driven complement to static structural retrieval, not a model-training procedure. Our evidence is limited to the evaluated benchmarks, library size, and backbones; scaling to substantially larger libraries and broader model families remains to be tested.

%% file: sections/appendix.tex
\appendix
\section*{Appendix}

The appendix has four parts: exact method and implementation details,
additional results and diagnostics, experimental protocol and reproducibility,
and scope and limitations.

\section{Method and Implementation Details}
\subsection{Algorithms}
\label{sec:appendix_alg}
Evolution is offline and runs in two phases, matching the code: Phase~A (Algorithm~\ref{alg:extract}) scans the collected traces and accumulates the execution signals. Phase~B (Algorithm~\ref{alg:update}) applies the three updates to the graph in dependency order. Inference (Algorithm~\ref{alg:retrieve}) runs the GoS retrieval procedure over the evolved graph.

\begin{algorithm}[H]
\caption{SE-GoS evolution (offline), Phase A: extract execution signals from traces.}
\label{alg:extract}
\begin{algorithmic}[1]
\REQUIRE Traces $\mathcal{D}=\{\mathcal{T}_t\}_{t=1}^{T}$, success threshold $r_{\text{success}}{=}0.9$, schema threshold $\zeta$, thresholds $\theta_{\text{obs}},\theta_{\text{avoid}}$, retraction threshold $\theta_{\text{retract}}{=}2$, node threshold $n_{\text{rank}}$
\STATE Initialize query-level counts $c^{\text{workflow}}_{uv}$, task-level counts $c^{\text{dependency}}_{uv},c^{\text{failed}}_{uv},c^{\text{success}}_{uv}$, surfaced-task indicator $z_t(v)$, and per-skill query map $Q_s\leftarrow\{s\mapsto\emptyset:s\in V\}$
\FOR{$t = 1, \dots, T$}
   \FORALL{queries $q$ recorded for task $t$}
      \STATE Record $z_t(v)=1$ iff $v$ appears in the retrieved top-$N$ for at least one query of task $t$
      \IF{$r_t\ge r_{\text{success}}$}
         \FORALL{lexical seeds $u$ of $q$, used $v\in U_t$, $u\neq v$} \STATE $c^{\text{workflow}}_{uv}\leftarrow c^{\text{workflow}}_{uv}+1$ \ENDFOR
      \ENDIF
   \ENDFOR
   \FORALL{ordered used pairs $u$ before $v$ in $U_t^{\mathrm{ord}}$, $\text{schema}(u,v)\ge\zeta$, $r_t\ge r_{\text{success}}$} \STATE $c^{\text{dependency}}_{uv}\leftarrow c^{\text{dependency}}_{uv}+1$ \ENDFOR
   \FORALL{trials $t$, pairs $\{u,v\}\subseteq U_t$} \STATE $c^{\text{failed}}_{uv} \leftarrow c^{\text{failed}}_{uv} + \mathbb{I}[r_t<r_{\text{success}}]$;\ $c^{\text{success}}_{uv} \leftarrow c^{\text{success}}_{uv} + \mathbb{I}[r_t\ge r_{\text{success}}]$ \ENDFOR
   \FORALL{failed trials $t$, queries $q$ of $t$, and $s\in U_t$ with $\text{rank}(s\mid q)>n_{\text{rank}}$} \STATE $Q_s\leftarrow Q_s\cup\{q\}$ \ENDFOR
\ENDFOR
\RETURN signal counts $(c^{\text{workflow}}, c^{\text{dependency}}, c^{\text{failed}}, c^{\text{success}}, Q_s)$
\end{algorithmic}
\end{algorithm}

\begin{algorithm}[H]
\caption{SE-GoS evolution (offline), Phase B: apply the topology, edge, and node updates.}
\label{alg:update}
\begin{algorithmic}[1]
\REQUIRE Static graph $G_0=(V,E_0,w_0,\phi)$, Phase-A signal counts, hyperparameters $\eta,\theta_{\text{obs}},\theta_{\text{avoid}},\theta_{\text{retract}}{=}2,n_{\text{rank}},p$
\STATE $E \leftarrow E_0$; $w \leftarrow w_0$ \textcolor{gray}{(start from the static graph)}
\STATE \textcolor{gray}{// (B.1) Topology update: induce workflow / dependency / avoid edges}
\FORALL{pairs $(u,v)$ with $c^{\text{workflow}}_{uv} \ge 1$} \STATE $E \leftarrow E \cup \{(u,v)\}$, type $\text{workflow}$, weight $w_{\text{workflow}}(u\to v)$ \ENDFOR
\FORALL{ordered pairs $(u,v)$ with $c^{\text{dependency}}_{uv} \ge 1$} \STATE $E \leftarrow E \cup \{(u,v)\}$, type $\text{dependency}$, weight $w_{\text{dependency}}(u\to v)$ \ENDFOR
\FORALL{pairs $(u,v)$ with $c^{\text{failed}}_{uv} \ge \theta_{\text{avoid}}$ and $c^{\text{success}}_{uv} = 0$} \STATE $E \leftarrow E \cup \{(u,v)\}$, type $\text{avoid}$, weight $0$ \ENDFOR
\FORALL{existing avoid edges $(u,v)$ with $c^{\text{success}}_{\{u,v\}}\ge\theta_{\text{retract}}$} \STATE $E\leftarrow E\setminus\{(u,v)\}$ \ENDFOR
\STATE \textcolor{gray}{// (B.2) L2 soft attenuation: each distinct task contributes at most once}
\FORALL{$v$ with $\sum_t z_t(v)\ge\theta_{\text{obs}}$ and $\sum_t\mathbb{I}[v\in U_t]=0$} \STATE $w(u\to v)\leftarrow0.5w(u\to v)$ for all semantic $u\to v$; retain edges, no floor \ENDFOR
\STATE \textcolor{gray}{// (B.3) L2 edge reinforcement (Eq.~\ref{eq:reinforce}); exclude avoid edges}
\FOR{$t$ with $r_t > 0$}
   \FORALL{used $v \in U_t$, edges $(u,v) \in E$ with $\phi(u,v)\neq\text{avoid}$} \STATE $w(u \to v) \leftarrow w(u \to v) + \eta \cdot r_t$ \ENDFOR
\ENDFOR
\STATE \textcolor{gray}{// (B.4) Node update (single round; no-eviction guard on)}
\FORALL{$s$ with $Q_s \neq \emptyset$} \STATE $d_s^{\star} \leftarrow \operatorname*{arg\,max}^{\mathrm{lex}}_{d \in \operatorname{Cands}(s)} (H_s(d),-R_s(d))$ subject to top-$K$ no-eviction guard; exact ties keep first candidate \hfill\textcolor{gray}{(Eq.~\ref{eq:desc})} \ENDFOR
\RETURN Evolved graph $G=(V,E,w,\phi)$ and descriptions $\{d_s^{\star}\}$
\end{algorithmic}
\end{algorithm}

\begin{algorithm}[H]
\caption{SE-GoS retrieval (inference).}
\label{alg:retrieve}
\begin{algorithmic}[1]
\REQUIRE Query $q$, evolved graph $G=(V,E,w,\phi)$, evolved descriptions $\{d_s^{\star}\}$, budget $\tau$
\STATE Compute the seed distribution $\mathbf{p}$ over lexical seed scores, using evolved descriptions
\STATE Build transition operator $T$ from evolved weights $w$ \hfill (Eq.~\ref{eq:transition})
\STATE Run reverse-aware PPR to convergence, $\mathbf{s}^{\star}$ \hfill (Eq.~\ref{eq:ppr})
\STATE Rank skills by $\mathbf{s}^{\star}$ and hydrate the bounded bundle
\RETURN Bounded execution bundle $B(q)$
\end{algorithmic}
\end{algorithm}

\subsection{Hyperparameters}
\label{sec:appendix_hyper}
SE-GoS uses a deterministic semantic-only cold-start graph by design. GoS can
construct additional typed relations when an offline LLM validation pass is
available, but SE-GoS omits that prior so that workflow, dependency, and avoid
relations are attributable to execution traces. Retrieval uses lexical seeding
and does not require an embedding service.

Table~\ref{tab:hyperparams} lists the SE-GoS hyperparameters and their default values. The retrieval-side hyperparameters (PPR restart, relation weights, budgets) are the harness defaults and are held fixed across benchmarks and library sizes so that any difference from the static baseline is attributable to evolution.

\begin{table}[H]
\centering
\footnotesize
\caption{SE-GoS hyperparameters; defaults used in all experiments.}
\label{tab:hyperparams}
\begin{tabular}{@{}p{2.6cm}p{4.4cm}p{4.6cm}@{}}
\toprule
Hyperparameter & Value & Role \\
\midrule
PPR restart $\alpha$ & 0.2 & Teleport probability (Eq.~\ref{eq:ppr}). \\
Reverse weights $\gamma_r$ & dependency 1.0, workflow 0.5, semantic 0.2 & Reverse-traversal strength (Eq.~\ref{eq:transition}). \\
Lexical seeding & weighted token overlap on name, capability, desc., tags, tooling, I/O, examples, entrypoints & Seed scores $p_i \propto s_i^{\text{lexical}}$; no embedding. \\
Top-$N$ / seed-$K$ / budgets & 5 / 4 / 1{,}800+9{,}000 chars & Bundle composition. \\
\midrule
Reinforcement rate $\eta$ & 0.1 & Hebbian step (Eq.~\ref{eq:reinforce}). \\
Induction weight & $0.6 + 0.05(c^{\text{workflow}}_{uv}-1)$, cap $0.9$ & Workflow-edge weight; count is query-level (Section~\ref{sec:evolution}). \\
Observation threshold $\theta_{\text{obs}}$ & 2 distinct tasks & Surface count before attenuating a never-used head; multiple queries in one task count once. \\
Semantic attenuation factor & 0.5 & One-time L2 scaling of incoming semantic weights; no edge deletion or floor. \\
Paraphrase count $p$ & 2 & Paraphrase variants per edited description. \\
Node-update threshold $n_{\text{rank}}$ & 3 & A used skill is targeted if ranked worse than this. \\
Node-update miss source & \texttt{failed} (default) & Queries from trials with reward $< r_{\text{success}}$. \\
Success threshold $r_{\text{success}}$ & 0.9 & Success cut for topology/node miss filters. \\
Avoid threshold $\theta_{\text{avoid}}$ & 1 & Failed co-occurrences (none successful) for avoid edge. \\
Avoid retraction threshold $\theta_{\text{retract}}$ & 2 distinct successful tasks & Successful co-use required to retract an existing avoid edge (L1). \\
Dependency threshold $\zeta$ & 0.6 & I/O schema overlap for dep certification. \\
No-eviction guard (node update) & \texttt{on} (default) & Across all recorded queries, preserve top-$K$ skills that were used in any recorded training trial. \\
Description edit budget & $\le$ 50 tokens & Cap on description edits. \\
\bottomrule
\end{tabular}
\end{table}

\subsection{Trace Extraction}
The used-skill set $U_t$ is extracted from the trajectory by scanning the agent's tool calls (shell, file-read, and code-execution calls) for references to the hydrated skill source paths returned in bundle $B_t$. The directory name is mapped to the normalized skill node via the source-path field of the skill record. $U_t$ is treated as a set for co-use statistics; dependency induction separately uses the recorded first-use order. Zero-reward trials contribute to failed-trial avoid evidence and to the L2 never-used attenuation statistics. Partial-reward trials contribute positive weight proportional to their reward (Eq.~\ref{eq:reinforce}).

\subsection{Node Update Prompts}
For each affected skill, the node update makes three LLM calls: a critic sees
the current description and all miss queries; an editor proposes one rewrite;
and a paraphraser proposes $p{=}2$ variants. The offline evaluator reruns the
same lexical-seed and PPR retrieval used at inference. It selects the candidate
maximizing hit count and then minimizing rank sum, commits only a candidate that
places the skill in the top-$n_{\text{rank}}$ for at least one miss query, and
enforces the no-eviction guard: across all recorded training queries, any skill used in a recorded training trial that currently appears in that query's top-$K$ must remain there. Exact score ties keep the first generated candidate. No agent run is involved.

\section{Additional Results and Diagnostics}

\subsection{Qualitative Analysis}
\label{sec:appendix_qual}
The evolution deltas themselves are inspectable. In the single evolution round, the topology update emits \emph{execution-grounded relations}: query-level workflow co-occurrences, dependency edges where ordered usage lines up with GoS's I/O-schema rule, and avoid edges from failed-trial co-use. The L1 update can retract a prior avoid relation only when at least two distinct successful task traces co-use the pair. L2 halves incoming semantic weights to skills surfaced by at least two distinct tasks but never used; it retains those edges and applies no floor. It also reinforces every existing incoming non-avoid edge to a used target on positive-reward trials. The node update changes graph-local retrieval descriptions for nodes that were used but ranked below the node-update threshold (Section~\ref{sec:method}). Reported counts in the results tables are from the evaluated run; updated method settings will be remeasured separately. Deltas record added and retracted edges and weight changes and can be inspected per edge.

\subsection{Configuration Rationale}
\label{sec:appendix_config}
This appendix gives the full argument behind the two configuration choices that the main text states tersely (Section~\ref{sec:experiment}).

\paragraph{Why a token-overlap semantic substrate.}
GoS builds its semantic neighbours from an embedding $k$NN index; SE-GoS's semantic edges are instead computed by signature-token Jaccard overlap. The computation is deterministic and model-free, so the base graph is built from raw text with no embedding service and stays recomputable after description edits, the indirect path by which semantic edges evolve (Section~\ref{sec:method}). An offline replay of the 87 task queries against both constructions shows they are retrieval-equivalent for the quantity we care about, whether the skills the agent later uses are surfaced: hit@5 and mean best rank are $70/73$ ($0.959$) and $1.19$ for both the official embedding-$k$NN base and the token-overlap base, while the token-overlap graph has a more uniform degree distribution (maximum degree $98$ on the released $k{=}1$ substrate, against $484$ for the embedding-$k$NN base), so PPR is less prone to hub collapse. Embedding-based edges would additionally tie the graph to a precomputed index that cannot be extended once descriptions change, since the deployment we target carries no embedding service.

\paragraph{Why no typed $\{\text{dependency}, \text{workflow}, \text{semantic}, \text{alternative}\}$ edge set.}
Reconstructing a typed edge set needs an LLM validation pass over a candidate pool built from per-skill I/O metadata, which the deployment SE-GoS targets does not assume. This is a design choice rather than a limitation: GoS's candidate pool has a lexical channel and an I/O-index channel that need no embedding service, so the typed graph is constructible without one, and SE-GoS's contribution is to obtain the structure from execution instead. Our static graph therefore contains only semantic edges, which is exactly the setting SE-GoS targets, and the missing structure is then supplied from execution feedback.

\paragraph{Why lexical-only seeding.}
We use the lexical-seed reverse-PPR condition supplied by GoS's SkillsBench evaluation harness. It requires no embedding service and keeps the seed and propagation configuration fixed between the static and evolved graphs; consequently, any difference is due to the evolved graph fields rather than a changed retrieval condition.

\subsection{Design Rationale (full arguments)}
\label{sec:appendix_design}
This appendix carries the design-rationale arguments in full.

\paragraph{Training-free, deterministic, auditable.}
The only consumer of the evolved graph beyond scoring is bundle composition, which drops co-loading avoid pairs when that channel is exercised (Section~\ref{sec:evolution}). Every other update reaches the retriever purely through the graph's weights, edges, and descriptions. Whether the evolved graph also lowers the input tokens actually consumed is evaluated in Section~\ref{sec:experiment}. Read together, the edge weights are a compressed, non-parametric record of successful executions that travels with the library and is shared across agents.

\paragraph{Prior versus operator.}
A \emph{prior} is a belief installed before evidence exists: GoS's relation validator and SkillDAG's pair classifier both judge relatedness from documents, having seen no execution. An \emph{operator} turns evidence into an update and may be a model: the node update's critic and editor act on a query that really used the skill and on retrieval evidence that really ranked it too low, and no candidate description is committed unless it demonstrably lifts the skill's rank on such a query. Had the structural channel carried an LLM's prior, the $2^3$ factorial could not have separated the gain due to evolution from the gain due to that prior, because the baseline itself would already contain it.

\paragraph{Why topology retraction is selective and attenuation is soft.}
L1 retracts only an avoid edge, and only when at least two distinct successful task traces later co-use that pair. This is direct counterevidence to the relation's failure-side definition; mere non-use is not. L2 handles weaker evidence by halving, rather than deleting, incoming semantic weights for repeatedly surfaced but never-used skills. Retaining the edge preserves the sparse substrate's connectivity and leaves a route for future evidence to reinforce it; no arbitrary weight floor is needed. This separates relation-level topology changes (induction and evidence-backed avoid retraction) from continuous semantic-weight attenuation. GoS also removes some edges during graph maintenance and deduplication, while SkillDAG permits explicit agent-proposed removal/retyping; neither is the same as deleting edges solely because they were not observed in a finite execution window~\citep{gos,skilldag}.

\paragraph{Overfitting and deployment windows.}
Experience-aware updating can overfit to traced tasks. The multi-round curve of Section~\ref{sec:multiround} shows that repeated updating plateaus at round~2 and reverses by round~3. One reading is that later rounds consume traces produced on the evolved graph, so the edges they add encode the previous round's ranking rather than new execution evidence. Production deployments can run evolution in windows, retaining the window-level statistics the topology and node updates require. We measure the single-round and multi-round regimes directly (Section~\ref{sec:multiround}) and do not characterize windowed deployment beyond that.

\paragraph{Confidence-weighted interpolation (deployment safeguard).}
In long-horizon deployment, where evolution repeats over many rounds and per-edge evidence counts stay small, a newly added or rarely observed edge should not immediately dominate a skill graph. We therefore blend the evolved weight with the static prior $w_0(e)$ set during graph construction,
\begin{equation}
w(e) = \big(1 - \lambda_c(e)\big)\, w_0(e) + \lambda_c(e)\, \tilde{w}(e),
\label{eq:conf}
\end{equation}
where $\tilde{w}(e)$ is the raw accumulated weight, $n_e$ is the number of informing trials for edge $e$, and
\begin{equation}
\lambda_c(e) = \frac{n_e}{n_e + n_0}
\label{eq:confweight}
\end{equation}
is a confidence weight with prior strength $n_0$. This interpolation is a deployment safeguard, not instantiated by the evaluated single-round implementation; a deployment using it must specify $n_0$ during calibration. For edges with no or few observations, $\lambda_c(e) \approx 0$ and retrieval behaves like the static prior. As evidence accumulates, the graph is increasingly driven by experience. In the reported single-round protocol, raw accumulated weights are used directly and this interpolation is not exercised.

\paragraph{Why graph-structured updating beats graph-free adaptation.}
A graph-free baseline weights each skill by its own experience score (success rate or average reward), treating skills as independent. SE-GoS instead updates \emph{edges}: when skill $u$ leads to a successful use of $v$, every future query that retrieves $u$ gains a path to $v$, a skill-to-skill transfer that a per-skill baseline, lacking any notion of relatedness, cannot express. The graph, not the experience signal alone, therefore carries the gain.

\paragraph{Completeness and update ordering.}
A skill graph has four components, $G=(V,E,w,\phi)$ (Eq.~\ref{eq:graph}). The three updates act on exactly the three \emph{mutable} ones, while the node set $V$ is fixed (SE-GoS never creates or deprecates skills). Within a round, updates run in dependency order: topology first, edge weights second as a refinement of the settled edge set, and the node update last because seed scores are computed from descriptions. In our lexical-only configuration, revised descriptions feed directly into the next seed's token overlap with no re-indexing step.

\paragraph{Why execution induces workflow, dependency, and avoid relations, and not alternative.}
The topology update emits three of GoS's four relations, a boundary on what execution \emph{identifies} rather than on what the machinery accepts, since the transition operator dispatches on the relation label by lookup. \emph{Workflow} is witnessed at query level: each recorded query's lexical seeds are paired with the skills used on that successful task; multiple queries in one task may witness the same ordered pair multiple times. This is co-occurrence evidence, not proof that retrieving $u$ caused use of $v$. \emph{Dependency} uses first-use order and GoS's I/O-schema rule. \emph{Avoid} is based on skills actually co-used on failed trials (not every skill merely loaded in the bundle), with successful co-use serving as counterevidence. \emph{Alternative} remains counterfactual and is not induced. The cold-start semantic graph remains in place, with L2 only attenuating semantic weights on repeatedly surfaced, never-used heads. The update issues no LLM calls. Recovering alternative without an LLM prior remains future work.

\paragraph{Why the dependency relation is induced but does not fire here.}
Dependency is the strictest of the three rules, and it is worth stating what that costs. A dependency edge needs a successful trial (reward $\ge 0.9$), two skills used in order, \emph{and} an I/O-schema overlap of at least $\zeta{=}0.6$ between the producer's outputs and the consumer's inputs. On the $k{=}1$ full-benchmark traces the first two conditions leave $68$ ordered pairs drawn from $22$ successful trials out of $87$. Of those pairs only $26$ carry I/O fields on both sides at all, and the largest overlap any of them attains is $0.500$, below $\zeta$. No dependency edge is induced, so the structural gain reported in this paper comes from the workflow edges the update adds (avoid edges carry no diffusion mass and only prune at composition), with the edge-weight and description updates on top of them.

The binding constraint is the skill library rather than the traces: the released $1{,}000$-skill library carries \texttt{inputs} for $332$ skills and \texttt{outputs} for $337$, so roughly two thirds of the library cannot take part in a schema match. This also delimits the ``structure from execution'' claim, and we prefer to draw the line explicitly rather than let it be inferred. Co-occurrence structure is observable in traces (retrieving $u$ and then using $v$ is witnessed directly, as is a pair that fails together), which is why workflow and avoid both fire. Mechanism is not: that $u$ produces an artifact $v$ consumes leaves no trace in the trajectory, which shows only that $u$ came first, and order by itself does not separate a prerequisite from a convention. GoS resolves this by reading the mechanism off the skill documents. We keep its rule and its threshold, so where the documents are silent the relation stays uninduced.

We did not lower $\zeta$ to obtain edges. At $\zeta{=}0.5$ eight pairs would qualify, but the threshold is GoS's, and moving it to manufacture a handful of edges is not a claim we could defend at this sample size. Two routes would make the relation live: recovering I/O fields from skill content, or mining writer$\to$reader data flow from the trajectories instead of reading schema. We probed the second: across the $87$ task trajectories it yields $12$ candidate pairs, none of which the schema channel corroborates, and the pairs that recur turn out to be sibling calls within a single task rather than prerequisites. Both remain future work.

\subsection{How Far the Avoid Channel Fires}
\label{sec:appendix_avoid}
With the default $\theta_{\text{avoid}}{=}1$, the update proposes $51$ failed-only
co-occurrences; the non-contradiction invariant withholds $24$, leaving $27$
avoid edges. Avoid edges have zero diffusion weight: they change co-loading at
bundle composition rather than PPR relevance scores.

\section{Experimental Protocol and Reproducibility}

\subsection{Benchmark, Backbone, and Accounting Details}
\label{sec:appendix_setup}
This subsection carries the setup detail that Section~\ref{sec:experiment} states only in summary.

\paragraph{Why the default backbone is a flash model.}
A smaller, cheaper backbone leaves more of the task difficulty to be resolved by the harness, which is the regime in which a retrieval-graph improvement is most visible and the configuration a cost-conscious deployment would run.

\paragraph{Task partitioning.}
The 87 tasks are \emph{partitioned}, never subsampled: every task is used in every regime, and Section~\ref{sec:protocol} states for each regime which subset a reported reward averages over.
For ALFWorld, we use the complete official \texttt{valid\_seen} split of 140 episodes, with no subsampling. This matches GoS's reported full 140-episode ALFWorld evaluation and SkillDAG's 140-episode \texttt{valid\_seen} evaluation; SkillDAG separately uses 420 episodes for its online-editing training phase. Thus, the $140\times2$ count in Table~\ref{tab:main} means two attempts on each episode in the full \texttt{valid\_seen} split, not a resampling to match SkillsBench's 87 tasks.

\paragraph{Attempt accounting.}
Every cell is measured at its full $87\times2=174$ attempt budget. For evolution, the parser keeps the final attempt for each task, so all three updates see $87$ per-task traces. The released code can pool attempts when \texttt{SEGOS\_MERGE\_ATTEMPTS=1}, but that option is off for the reported results.

\paragraph{SkillsBench reward.}
For each generated task copy, the harness runs the upstream pytest suite and records a CTRF report. Our local verifier reward is the number of collected test cases marked \texttt{passed} divided by the total number collected; failed, skipped, pending, and other statuses remain in the denominator and do not count as passes. This wrapper changes only reward aggregation, not the task instructions or test cases; the source task packages remain unchanged. The fractional reward is the score used for SkillsBench runs performed in this work, while previously published baseline entries retain their source-reported values.

\paragraph{Agent harness.}
All three backbones are driven through the same Codex-CLI agent harness, and the wrapper is the only component that differs between the model blocks.

\paragraph{The $2^3$ design.}
Deltas are computed once from the traces and combined into an eight-cell $2^3$ design (static plus seven evolved cells), reported in Section~\ref{sec:ablation}. Within a round the updates run in dependency order, topology first, then edge weights, then node descriptions, for the reasons given in Appendix~\ref{sec:appendix_design}.

\subsection{Protocol Comparison with Self-Evolving Baselines}
\label{sec:appendix_protocol}
SkillDAG~\citep{skilldag}, the closest concurrent self-evolving skill-graph system, evaluates SkillsBench \emph{in-domain}: the graph is edited online during execution (a propose-edge/edit-edge pair at episode time), reward is measured as the agent and the evolving graph work together on the same tasks, and a cold-vs-edited replay of the same queries isolates retrieval mechanics (Ret@K/MRR) before any downstream execution. Its generalization claim rests on an ALFWorld train/test split: 420 in-domain training episodes with edits enabled, and 140 held-out test episodes evaluated on both the cold-start and the training-produced graph. Its cold-start graph is also LLM-dependent (a HyDE-style \texttt{e\_needs} embedding plus an LLM pair classifier), and its own premise is that any cold-start graph is necessarily incomplete until execution feedback arrives. SE-GoS shares this premise but removes the LLM dependency: it begins from a deterministic semantic-only similarity graph and reconstructs the missing execution structure with fixed rules. SkillGraph-RL~\citep{skillgraph-rl}, the training-based alternative, likewise trains on in-domain domains (NQ, HotpotQA) and evaluates on disjoint held-out/unseen domains.

Our SkillsBench evaluation follows SkillDAG's deployment-utility convention (all $87$ tasks generate the traces and all $87$ are re-measured), so the headline number is directly comparable to its in-domain number. Because SE-GoS evolves offline in one batch, updates fitted to the same queries that are later scored can reflect memorization rather than transferable structure. We separate the two claims by also reporting a held-out evaluation in which the graph is evolved on one task subset and measured on a disjoint one (Section~\ref{sec:heldout}). We also borrow SkillDAG's cold-vs-edited idea as a complementary retrieval-recall diagnostic: comparing whether the evolved graph surfaces the skills the agent actually uses more often than the static graph. This diagnostic is reported with the evolution results rather than used as a headline claim.

\subsection{Benchmark-Fidelity Audit}
\label{sec:appendix_fidelity}
Our task packages descend from the upstream SkillsBench repository~\citep{li2026skillsbench}, at the snapshot immediately before its schema migration (commit \texttt{d75b2187}, 2026-06-14), i.e.\ the v1.1 \texttt{task.toml} layout, prior to the repackaging of tasks into native \texttt{task.md} frontmatter. Comparing each of our 87 task packages against that snapshot file-by-file:
\begin{itemize}
\item \texttt{instruction.md} (the prompt given to the agent): $0$ of $87$ differ;
\item \texttt{tests/} (the reward verifier, including \texttt{test.sh}, \texttt{test\_outputs.py}, \texttt{score\_outputs.py}, and \texttt{build.sh}): $0$ of $87$ differ;
\item \texttt{task.toml} (timeouts, CPU/memory/storage limits, network policy): $0$ of $87$ differ.
\end{itemize}
The three components that determine task difficulty and reward are therefore byte-identical to upstream v1.1: no instruction, verifier threshold, timeout, or resource setting was relaxed, so none of the gap between SE-GoS and the published baselines can be attributed to a weakened task configuration.

The only differences from that snapshot lie in the \emph{Docker build layer} of seven tasks, and are local accommodations that do not change the task logic or the verifier:
\begin{itemize}
\item \textbf{Four tasks} carry a \texttt{predownload/} directory (\texttt{seismic-\allowbreak phase-\allowbreak picking}, \texttt{earthquake-\allowbreak phase-\allowbreak association}, \texttt{fix-\allowbreak druid-\allowbreak loophole-\allowbreak cve}, and \texttt{python-\allowbreak scala-\allowbreak translation}). The upstream Dockerfiles fetch the same artifacts at build time: seisbench model weights (with retry), the Apache Druid 0.20.0 tarball plus shallow source clone, and the Scala 2.13.12 distribution. Our build copies them from a local cache because the egress gateway blocks those hosts. The resulting images are equivalent.
\item \texttt{fix-\allowbreak build-\allowbreak google-\allowbreak auto} deletes unused Oracle JDKs inherited from the bugswarm base image to avoid a licensing compliance alert.
\item \textbf{Two web tasks} (\texttt{fix-visual-stability}, \texttt{react-performance-debugging}) pin \texttt{next@14.2.10} rather than upstream's \texttt{14.0.4}/\texttt{14.2.0}. This is the single true dependency-version difference. It does not touch the verifier. Reverting these two packages to the upstream pins and re-measuring is a two-task check if exact reproducibility of the images is ever required.
\end{itemize}
One oracle script (\texttt{pptx-reference-formatting}/\texttt{solution}/\texttt{solve.sh}) also differs in an error-message string only. Oracles never determine reward.

\section{Scope and Limitations}
\label{sec:appendix_limitations}
Four constraints bound the claims above.

\noindent\textbf{Single-round headline.} Repeating the update plateaus at round~2 (59.8\%) and reverses by round~3 (54.0\%, Section~\ref{sec:multiround}), so the evolved graph is reported after a single round and the protocol does not reach finer-grained reward signals or longer horizons. The reversal bounds the reported regime rather than the updates themselves: the edges induced after the first round stop paying for themselves, and Appendix~\ref{sec:appendix_design} gives the mechanism we propose for that.

\noindent\textbf{Used-set extraction.} The weight and topology updates read the used-skill set $U_t$ recovered from tool calls, so indirect skill use can undercount evidence. The rule errs toward omission: an unrecovered use contributes no evidence rather than a wrong one, which weakens the update instead of corrupting it.

\noindent\textbf{Evaluation scope.} We evaluate SkillsBench at 1{,}000 skills and the ALFWorld dev split under the three model families of Table~\ref{tab:main}, so the numbers bound what SE-GoS achieves in these settings and on these backbones, and ALFWorld is close to saturation there, which is why the two follow-up studies run on SkillsBench alone. We make no claim about other backbones or about the 200/500/2{,}000-skill scales.

\noindent\textbf{Cold start and convergence.} The confidence-weighted interpolation (Eq.~\ref{eq:conf}) protects low-evidence edges, but cold-start and long-horizon convergence are only partially characterized, since we measure rounds 2 and 3 and stop there. Recovering I/O fields for libraries like the one we use, so that the dependency relation can fire, remains future work.